\documentclass[conference]{IEEEtran}
\usepackage{graphicx}
\usepackage{verbatim}
\usepackage{amsmath,  amssymb}
\usepackage{subfigure}
\usepackage{algorithm}
\usepackage{algorithmic}
\usepackage{color}
\usepackage{float}
\usepackage{helvet}

\begin{document}

\title{Feature Selection via Regularized Trees}

\author{\IEEEauthorblockN{Houtao Deng}
\IEEEauthorblockA{Intuit\\
Mountain View, California, USA\\
Email: hdeng3@asu.edu}
\and
\IEEEauthorblockN{George Runger}
\IEEEauthorblockA{School of Computing, Informatics \& Decision Systems Engineering\\
Arizona State university, Tempe, Arizona, USA\\
Email: george.runger@asu.edu}}

\maketitle

\begin{abstract}
We propose a tree regularization framework, which enables many tree models to perform feature selection efficiently. The key idea of the regularization framework is to penalize selecting a new feature for splitting when its gain (e.g. information gain) is similar to the features used in previous splits. The regularization framework is applied on random forest and boosted trees here, and can be easily applied to other tree models. Experimental studies show that the regularized trees can select high-quality feature subsets with regard to both strong and weak classifiers. Because tree models can naturally deal with categorical and numerical variables, missing values, different scales between variables, interactions and nonlinearities etc., the tree regularization framework provides an effective and efficient feature selection solution for many practical problems.
\end{abstract}


\newtheorem{Definition}{Definition}
\newtheorem{Lemma}{Lemma}

\section{Introduction}
In supervised learning, given a training data set consisting of $N$ instances, $M$ predictor variables ${X_1,X_2,...X_M}$ and the target variable $Y\in\{0,1,...C-1\}$, feature selection is commonly used to select a compact feature subset $F\subset\{X_1,X_2,...X_M\}$ without significant loss of the predictive information about $Y$. Feature selection methods play an important role in defying the curse of dimensionality, improving efficiency both in time and space, and facilitating interpretability \cite{guyon2003}.

We propose a tree regularization framework for feature selection in decision trees. The regularization framework avoids selecting a new feature for splitting the data in a tree node when that feature produces a similar gain (e.g. information gain) to features already selected, and thus produces a compact feature subset. The regularization framework only requires a single model to be built, and can be easily added to a wide range of tree-based models which use one feature for splitting data at a node. We implemented the regularization framework on random forest (RF) \cite{breiman2001} and boosted trees \cite{freund1996experiments}. Experiments demonstrate the effectiveness and efficiency of the two regularized tree ensembles. As tree models naturally handle categorical and numerical variables, missing values, different scales between variables, interactions and nonlinearities etc., the tree regularization framework provides an effective and efficient feature selection solution for many practical problems.

Section \ref{sec:background} describes related work and background. Section \ref{sec:motivation} presents the relationship between decision trees and the Max-Dependency scheme \cite{penghanchuan2006}. Section \ref{sec:reg} proposes the tree regularization framework, the regularized random forest (RRF) and the regularized boosted random trees (RBoost). Section \ref{sec:evaluation} establishes the evaluation criteria for feature selection. Section \ref{sec:experiment} demonstrates the effectiveness and efficiency of RRF and RBoost by extensive experiments. Section \ref{sec:conclusion} concludes this work.

\section{Related Work and Background}\label{sec:background}
\subsection{Related work}
Feature selection methods can be divided into filters, wrappers and embedded methods \cite{guyon2010model}. Filters select features based on criteria independent of any supervised learner \cite{hall2000,liuhuan2004}.  Therefore, the performance of filters may not be optimum for a chosen learner. Wrappers use a learner as a black box to evaluate the relative usefulness of a feature subset \cite{kohavi1997wrappers}. Wrappers search the best feature subset for a given supervised learner, however, wrappers tend to be computationally expensive \cite{liu2005toward}.

Instead of treating a learner as a black box, embedded methods select features using the information obtained from training a learner. A well-known example is SVM-RFE (support vector machine based on recursive feature elimination) \cite{guyon2002}. At each iteration, SVM-RFE eliminates the feature with the smallest weight obtained from a trained SVM. The RFE framework can be extended to classifiers able to provide variable importance scores, e.g. tree-based models \cite{diaz2006gene}. 
Also, decision trees such as C4.5 \cite{quinlan1993c4} are often used as embedded methods as they intrinsically perform feature selection at each node. Single tree models were used for feature selection \cite{frey2003}, however, the quality of the selected features may be limited because the accuracy of a single tree model may be limited. In contrast, tree ensembles, consisting of multiple trees are believed to be significantly more accurate than a single tree \cite{breiman2001}. However, the features extracted from a tree ensemble are usually more redundant than a single tree. Recently, \cite{tuv2009} proposed ACE (artificial contrasts with ensembles) to select a feature subset from tree ensembles. ACE selects a set of relevant features using a random forest \cite{breiman2001}, then eliminates redundant features using the surrogate concept \cite{breiman1983cart}. Also multiple iterations are used to uncover features of secondary effects. 

The wrappers and embedded methods introduced above require building multiple models, e.g. the RFE framework \cite{guyon2002} requires building potentially $O(M)$ models. Even at the expense of some acceptable loss in prediction performance, it is very desirable to develop feature selection methods that only require training a single model which may considerably reduce the training time \cite{guyon2010model}.
The tree regularization framework proposed here enables many types of decision tree models to perform feature subset selection by building the models only one time. Since tree models are popularly used for data mining, the tree regularization framework provides an effective and efficient solution for many practical problems.

\def\myWidth{2.2}
\begin{figure*}[!]
\centering
\subfigure[A decision tree may use both $X_1$ and $X_2$ to split. ]{
\includegraphics[width= \myWidth in]{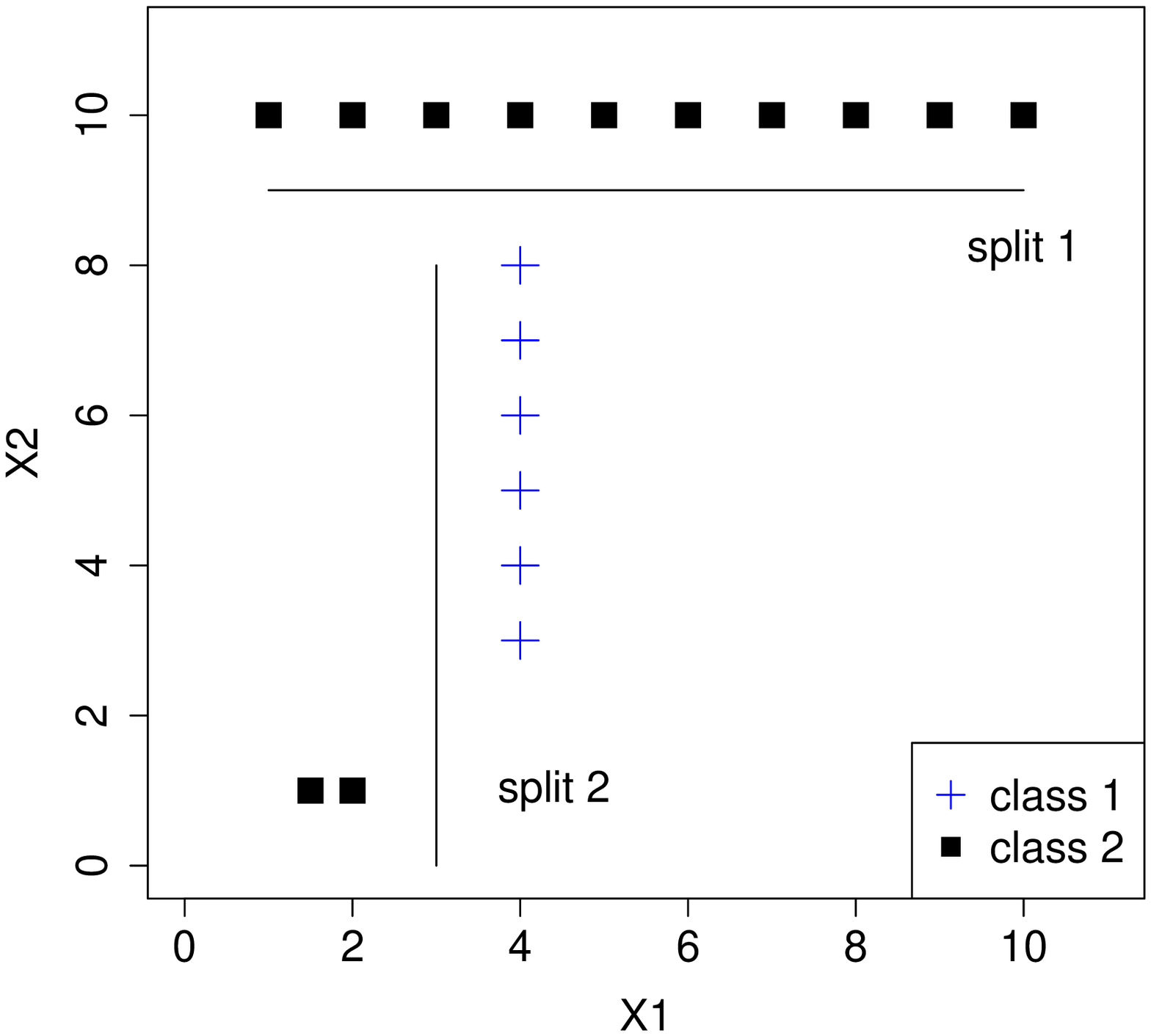} 
}
\subfigure[$X_2$ alone can perfectly separate the two classes.]{
\includegraphics[width= \myWidth in]{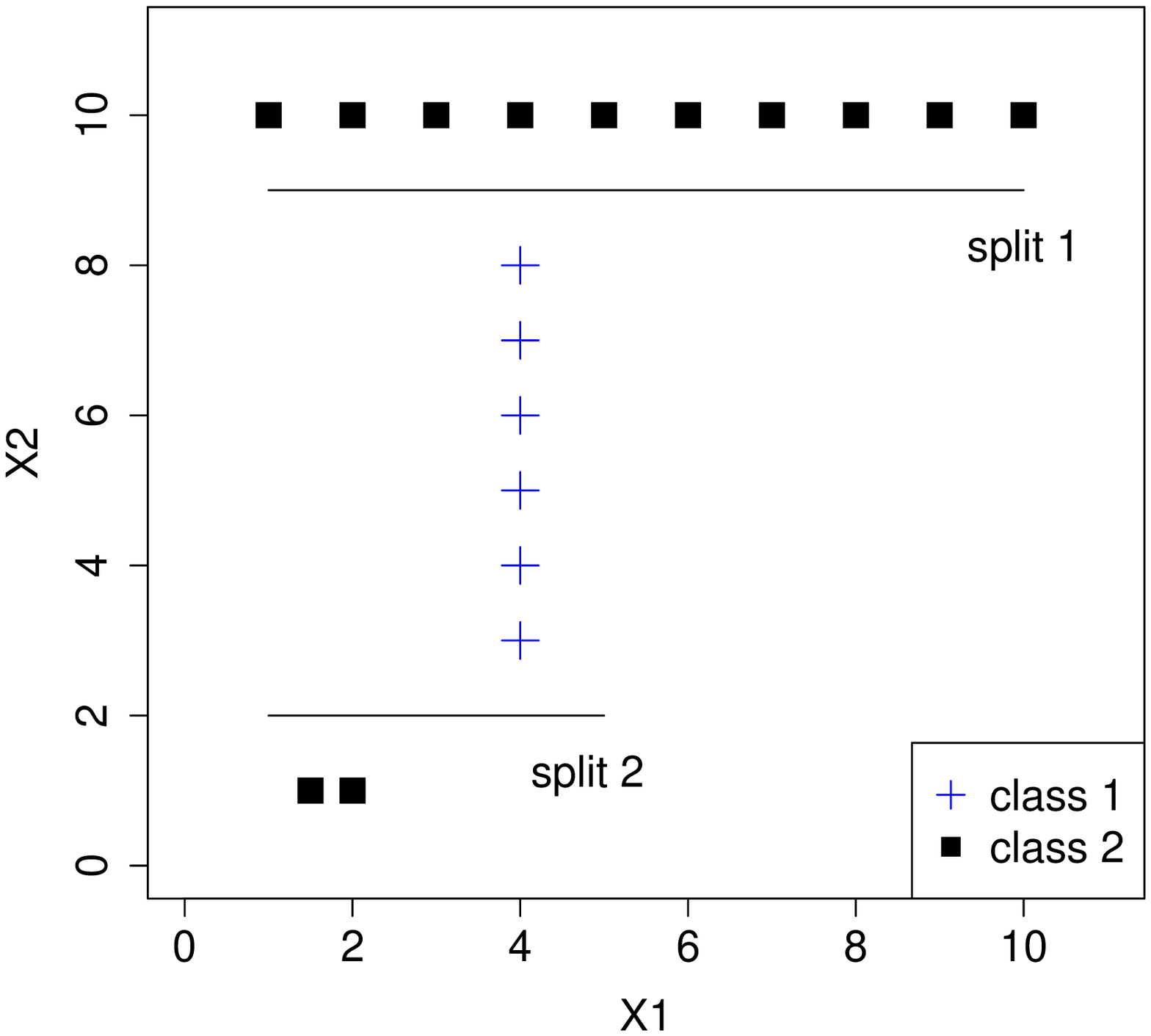} 
}

\caption[2]{An illustration of feature redundancy in decision trees. A decision tree may use both features to split, but $X_2$ alone can perfectly separate the two classes.\label{fig:split}}
\end{figure*}

\subsection{Information-theoretic measures and issues}\label{sec:inf}
Information-theoretic measures have been widely used for feature selection \cite{jakulin2003analyzing,fleuret2004fast,liuhuan2004,hall2000,penghanchuan2006}.
Entropy is an important concept in the information-theoretic criteria. The entropy of a categorical variable $A$ can be expressed in terms of prior probabilities:
$H(A)=-\sum_{a\in A}p(a)\log_2p(a)$.
The entropy of $A$ after observing another categorical variable $B$ is:
$H(A|B)=-\sum_{b\in B}p(b)\sum_{a\in A}p(a|b)\log_2p(a|b)$.
The increase in the amount of information about $A$ after observing $B$ is called the \textbf{mutual information} or, alternatively, \textbf{information gain} \cite{hall2000}:
\begin{equation}\label{equation:mutual}
I(A;B)=H(A)-H(A|B)
\end{equation}
$I(A;B)$ is symmetric, i.e. $I(A;B)=I(B;A)$, and models the degree of association between $A$ and $B$. Therefore, one can use $I(X_i;Y)$ to evaluate the relevancy of $X_i$ for predicting the class $Y$, and use $I(X_i;X_j)$ to evaluate the redundancy in a pair of predictor variables \cite{penghanchuan2006}. In addition, a measure called symmetric uncertainty: $SU(A;B)=2(H(A)-H(A|B))/(H(A)+H(B))$ is used in feature selection methods such as CFS (correlation-based feature selection) \cite{hall2000} and FCBF (fast correlation-based filter) \cite{liuhuan2004}.

Measures like $I(A;B)$ and $SU(A;B)$ capture only two-way relationships between variables and can not capture the relationship between two variables given other variables, e.g. $I(X_1;Y|X_2)$ \cite{jakulin2003analyzing, fleuret2004fast}. \cite{fleuret2004fast} illustrated this limitation using an exclusive OR example: $Y=XOR(X_1,X_2)$, in which neither $X_1$ nor $X_2$ individually is predictive, but $X_1$ and $X_2$ together can correctly determine $Y$. To this end, \cite{jakulin2003analyzing, fleuret2004fast} proposed measures which can capture three-way interactions. Still, a feature selection method capable of handling $n$-way interactions when $n>3$ is desirable \cite{jakulin2003analyzing}. However, it is computationally expensive to do so \cite{fleuret2004fast}.

\subsection{Tree-based models and issues}
Univariate decision trees such as C4.5 \cite{quinlan1993c4} or CART \cite{breiman1983cart} recursively split data into subsets. For many tree models, the feature used for splitting in a node is selected to optimize an information-theoretic measure such as information gain. 

A tree model is able to capture multi-way interactions between the splitting variables and potentially is a solution for the issue of the information-theoretic measures mentioned in Section \ref{sec:inf}. However, tree models have their own problems for selecting a non-redundant feature set. A decision tree selects a feature at each node by optimizing, commonly, an information-theoretic criterion and does not consider if the feature is redundant to the features selected in previous splits, which results in feature redundancy. The feature redundancy problem in tree models is illustrated in Figure \ref{fig:split}. For the two-class data shown in the figure, after splitting on $X_2$ (``split 1"), either $X_1$ or $X_2$ can separate the two classes (``split 2"). Therefore $\{X_2\}$ is the minimal feature set that can separate the two-class data. However, a decision tree may use $X_2$ for ``split 1" and $X_1$ for ``split 2" and thus introduce feature redundancy.

The redundancy problem becomes even more severe in tree ensembles which consist of multiple trees. To eliminate the feature redundancy in a tree model, some regularization is used here to penalize selecting a new feature similar to the ones selected in previous splits.

\def\myWidth{2.5}
\begin{figure*}[ht]
\centering
\subfigure[At each level, a decision tree can have different variables for splitting the nodes. \label{fig:MultilevelTree}]{
\includegraphics[width= \myWidth in]{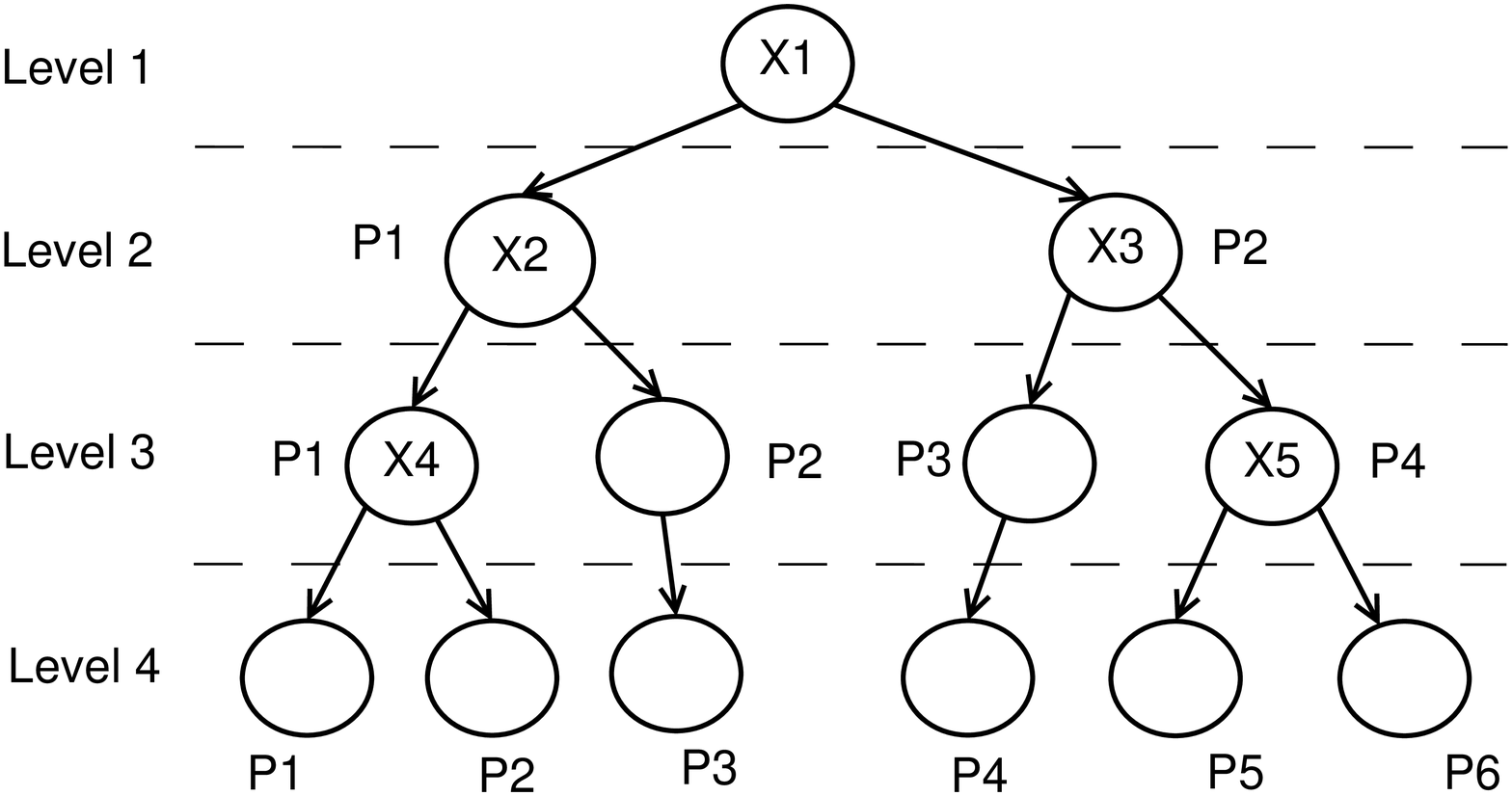} 
}
\subfigure[At each level, the MD scheme uses only one variable for splitting all the nodes. 
\label{fig:MultilevelMD}]{
\includegraphics[width= \myWidth in]{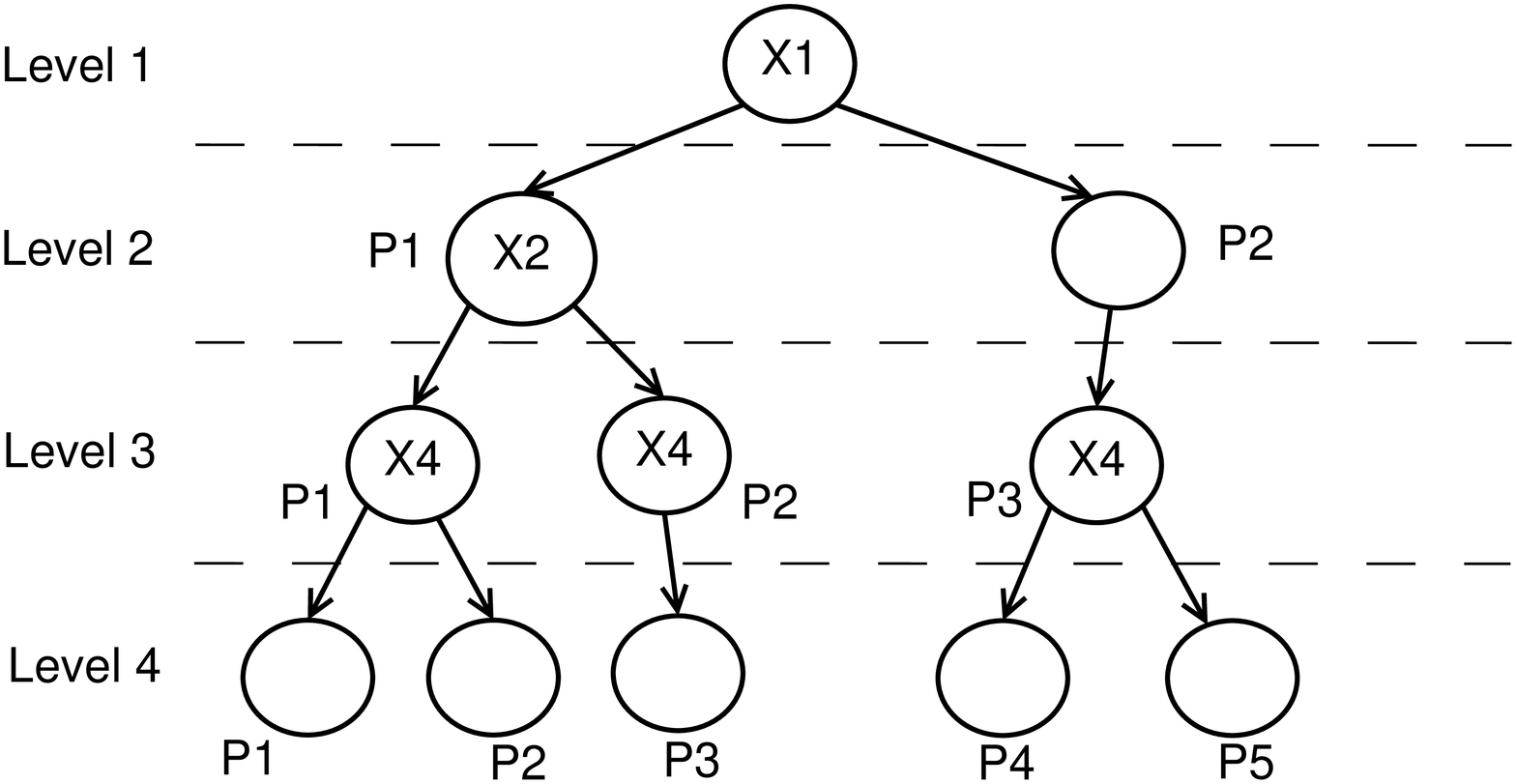} 
}
\caption[2]{Illustrations of a decision tree and the MD scheme in terms of a tree structure. A node having more than one child node is marked with the splitting variable. For a decision tree node that can not be split, we let all the instances in the node pass to its ``imaginary'' child node, to keep a form similar to the MD tree.\label{fig:treeMD}}
\end{figure*}

\begin{algorithm*}[ht]
\caption{{Feature selection via the regularized random tree model: $F=tree(data,F,\lambda)$, where $F$ is the feature subset selected by previous splits and is initialized to an empty set. Details not directly relevant to the regularization framework are omitted. Brief comments are provided after ``$//$". }} \label{alg:tree}
\begin{algorithmic}[1]
\STATE $gain^* = 0$
\STATE $count = 0$ // the number of new features tested
\FOR{$m=1:M$}\label{stepReg1} 
 \STATE{$gain_R(X_m)$=0}
  \STATE{\textbf{if} $X_m \in F$  \textbf{then} $gain_R(X_m) = gain(X_m)$ \textbf{end if} \label{step2} //calculate the $gain_R$ for all variables in $F$}

 \IF{$X_m \notin F$ and $count<\lceil \sqrt M \rceil$}
   \STATE{$gain_R(X_m) = \lambda \cdot gain(X_m)$ //{penalize using new features}}
   \STATE{$count$ = $count$+1}
 \ENDIF

 \STATE{\textbf{if} $gain_R(X_m) > gain^*$ \textbf{then} $gain^* = gain_R(X_m)$, $X^*=X_m$ \textbf{end if}}
\ENDFOR \label{stepReg2}

\STATE{\textbf{if} $gain^* = 0$ \textbf{then} make this node as a leaf and return $F$ \textbf{end if}}
\STATE {\textbf{if} $X^* \notin F$ \textbf{then} $F=\{F,X^*\}$} \textbf{end if} \label{stepSplit}
\STATE split $data$ into $\gamma$ child nodes by $X^*$: $data_1$, ...$data_\gamma$
\FOR{$g=1:\gamma$}\label{step3}
\STATE $F=tree(data_{g},F,\lambda)$
\ENDFOR
\STATE return $F$
\end{algorithmic}
\end{algorithm*}

\section{Relationship between decision trees and the Max-Dependency scheme}\label{sec:motivation}
The \textbf{conditional mutual information}, that is, the mutual information between two features $A$ and $B$ given a set of other features $C_1$, ...$C_p$, is defined as

\begin{align}\label{eq:condMutual}
 & I(A;B|C_1,...C_p)= \notag \\
 & \sum_{c_1 \in C_1}...\sum_{c_p \in C_p} w_{C_1=c_1,...C_p=c_p} I(A;B|C_1=c_1,...C_p=c_p)
\end{align}
where $w_{C_1=c_1,...C_p=c_p}$ is the ratio of the number of instances satisfying $\{C_1=c_1,...C_p=c_p\}$ to the total number of instances.

A first-order incremental feature selection scheme, referred to as the \textbf{Max-Dependency (MD)}\cite{penghanchuan2006} scheme, is defined as
\vspace{-5 pt}
\begin{equation}\label{eq:maxRel}
i = \arg\max_{m=1}^M I(X_m;Y|F(j-1)); F(j)= \{F(j-1),X_i\}
\end{equation}
where $j$ is the step number, $F(j)$ is the feature set selected in the first $j$ steps ($F(0)=\emptyset$), $i$ is the index of the feature selected at each step, $I(X_m;Y|F(j-1))$ is the mutual information between $X_m$ and $Y$ given the feature set $F(j-1)$.

Here we consider 
the relationship between the MD scheme and decision trees. Because Equation (\ref{eq:condMutual}) is limited to categorical variables, the analysis in this section is limited to categorical variables. We also assume the decision trees discussed in this section select the splitting variable by maximizing the information gain and split a non-leaf node into $K$ child nodes, where $K$ is the number of values of the splitting variable. However the tree regularization framework introduced later is not limited to such assumptions.

In a decision tree, a node can be located by its level (depth) $L_j$ and its position in that level. An example of a decision tree is shown in Figure \ref{fig:MultilevelTree}. The tree has four levels, and one to six nodes (positions) at each level. Note that in the figure, a tree node that is not split is not a leaf node. Instead, we let all the instances in the node pass to its ``imaginary'' child node, to keep a form similar to the MD tree structure introduced later. 

Also, let $S_{\nu}$ denote the set of feature-value pairs that define the path from the root node to node $\nu$. For example, for node P6 at level 4 in Figure \ref{fig:MultilevelTree}, $S_{\nu} = \{X_1 = x_1, X_3 = x_3, X_5= x_5\}$. For a decision tree node $\nu$, a variable $X_k$ is selected to maximize the information gain conditioned on $S_{\nu}$. That is,
\begin{equation}\label{eq:treeSelection}
k = \arg\max_{m=1}^M I(X_m;Y|S_{\nu})
\end{equation}

By viewing each step of the MD scheme as a level in a decision tree, the MD scheme can be expressed as a tree structure, referred to an MD tree. An example of an MD tree is shown in Figure \ref{fig:MultilevelMD}. Note in an MD tree, only one feature is selected at each level. Furthermore, for the MD tree, $X_k$ is selected at $L_j$ so that
\begin{equation}\label{eq:mdSelection}
k = \arg\max_{m=1}^M \sum_{\nu \in L_j} w_{\nu}*I(X_m;Y|S_{\nu})
\end{equation}
where $w_{\nu}$ is the ratio of the number of instances at node $\nu$ to the total number of training instances.

Note Equation (\ref{eq:treeSelection}) maximizes the conditional mutual information at each node, while Equation (\ref{eq:mdSelection}) maximizes a weighted sum of the conditional mutual information from all the nodes in the same level. Calculating Equation (\ref{eq:mdSelection}) is more computationally expensive than Equation (\ref{eq:treeSelection}). However, at each level $L_j$, an MD tree selects only one feature that adds the maximum non-redundant information to the selected features, while decision trees can select multiple features and there is no constraint on the redundancy of these features. 


\section{Regularized trees}\label{sec:reg}
We are now in a position to introduce the tree regularization framework which can be applied to many tree models which recursively split data based on a single feature at each node. Let $gain(X_j)$ be the evaluation measure calculated for feature $X_j$. Without loss of generality, assume the splitting feature at a tree node is selected by maximizing $gain(X_j)$ (e.g. information gain). Let $F$ be the feature set used in previous splits in a tree model. When the tree model is built, then $F$ becomes the final feature subset.


The idea of the tree regularization framework is to avoid selecting a new feature $X_j$, i.e., avoid features not belonging to $F$, unless $gain(X_j)$ is substantially larger than $\max_i(gain(X_i))$ for $X_i\in F$. To achieve this goal, we consider a penalty to $gain(X_j)$ for $X_j \notin F$. A new measure is calculated as
 \begin{equation}
gain_R(X_j) = \begin{cases}
 \lambda \cdot gain(X_j) & \text{$X_j \notin F$} \\
 gain(X_i) & \text{$X_j \in F$} \\
\end{cases}
\end{equation}
where $\lambda\in[0,1]$. Here $\lambda$ is called the coefficient. A smaller $\lambda$ produces a larger penalty to a feature not belonging to $F$. Using $gain_R(\cdot)$ for selecting the splitting feature at each tree node is called a \textbf{tree regularization framework}.
A tree model using the tree regularization framework is called a regularized tree model.
A regularized tree model sequentially adds new features to $F$ if those features provide substantially new predictive information about $Y$. The $F$ from a built regularized tree model is expected to contain a set of informative, but non-redundant features. Here $F$ provides the selected features directly, which has the advantage over a feature ranking method (e.g. SVM-RFE) in which a follow-up selection rule needs to be applied.


A similar penalized form to $gain_R(\cdot)$ was used for suppressing spurious interaction effects in the rules extracted from tree models \cite{friedmanpredictive2008}. The objective of \cite{friedmanpredictive2008} was different from the goal of a compact feature subset here.
Also, the regularization in \cite{friedmanpredictive2008} only reduced the redundancy in each path from the root node to a leaf node, but the features extracted from tree models using such a regularization \cite{friedmanpredictive2008} can still be redundant.  

Here we apply the regularization framework on the random tree model available at Weka \cite{weka}. The random tree randomly selects and tests $K$ variables out of $M$ variables at each node (here we use $K=\lceil\sqrt M\rceil$ which is commonly used for random forest \cite{breiman2001}), and recursively splits data using the information gain criterion.

The random tree using the regularization framework is called the regularized random tree algorithm which is shown in Algorithm \ref{alg:tree}. The algorithm focuses on illustrating the tree regularization framework and omits some details not directly relevant to the regularization framework. The regularized random tree differs from the original random tree in the following ways: 1) $gain_R(X_j)$ is used for selecting the splitting feature; 2) $gain_R$ of all variables belonging to $F$ are calculated, and the $gain_R$ of up to $\lceil\sqrt M\rceil$ randomly selected variables not belonging to $F$ are calculated. Consequently, to enter $F$ a variable needs to improve upon the gain of all the currently selected variables, even after its gain is penalized with $\lambda$.


\begin{algorithm}[h]
\caption{{Feature selection via the regularized tree ensemble: $F=ensemble(data,F,\lambda,nTree)$, where $F$ is feature subset selected by previous splits and is initialized to an empty set, $nTree$ is the number of regularized trees in the tree ensemble.}
 \label{alg:treeEnsemble}}
\begin{algorithmic}[1]
\FOR{iTree = 1:$nTree$}
 \STATE select $data_i$ from $data$ with some criterion, e.g. randomly select
 \STATE $F=tree(data_i,F,\lambda)$
\ENDFOR
\end{algorithmic}
\end{algorithm}

The tree regularization framework can be easily applied to a tree ensemble consisting of multiple single trees. The regularized tree ensemble algorithm is shown in Algorithm \ref{alg:treeEnsemble}. $F$ now represents the feature set used in previous splits not only from the current tree, but also from the previous built trees. Details not relevant to the regularization framework are omitted in Algorithm \ref{alg:treeEnsemble}. The computational complexity of a regularized tree ensemble with $nTree$ regularized trees is $nTree$ times the complexity of the single regularized tree algorithm. The simplicity of Algorithm \ref{alg:treeEnsemble} suggests the easiness of extending a single regularized tree to a regularized tree ensemble. Indeed, the regularization framework can be applied to many forms of tree ensembles such as bagged trees \cite{breimanbagging1996} and boosted trees \cite{freund1996experiments}.
In the experiments, we applied the regularization framework to bagged random trees, referred to as random forest (RF) \cite{breiman2001}, and boosted random trees. The regularized versions are called the \textbf{regularized random forest} (\textbf{RRF}) and \textbf{regularized boosted random trees} (\textbf{RBoost}).


\section{Evaluation criteria for feature selection}\label{sec:evaluation}
A feature selection evaluation criterion is needed to measure the performance of a feature selection method. Theoretically, the optimal feature subset should be a minimal feature set without loss of predictive information and can be formulated as a Markov blanket of $Y$ ($MB(Y)$) \cite{Koller1996,aliferis2010local}. The Markov blanket can be defined as \cite{aliferis2010local}:
\begin{Definition}\label{def:markov}
\textbf{Markov blanket of Y}:
A set $MB(Y)$ is a minimal set of features with the following property. For each feature subset $f$ with no intersection with $MB(Y)$, $Y\perp f|MB(Y)$. That is, $Y$ and $f$ are conditionally independent given $MB(Y)$. In \cite{pearl1988}, this terminology is called the Markov Boundary.
\end{Definition}

In practice, the ground-truth $MB(Y)$ is usually unknown and the evaluation criterion of feature selection is commonly associated with the expected loss of a classifier model, referred to as the empirical criterion here (similar to the definition of ``feature selection problem" \cite{aliferis2010local}):

\begin{Definition}\label{def:emperr}
\textbf{Empirical criterion}:
Given a set of training instances of instantiations of feature set $X$ drawn from distribution $D$, a classifier induction algorithm $C$, and a loss function $L$, find the smallest subset of variables $F\subseteq X$ such that $F$ minimizes the expected loss $L(C,D)$ in distribution $D$.
\end{Definition}

The expected loss $L(C,D)$ is commonly measured by classification generalization error. According to Definition \ref{def:emperr}, to evaluate two feature subsets, the subset with a smaller generalization error is preferred. With similar errors, then the smaller feature subset is preferred. 

Both evaluation criteria prefer a feature subset with less loss of predictive information. However, the theoretical criterion (Definition \ref{def:markov}) does not depend on a particular classifier, while the empirical criterion (Definition \ref{def:emperr}) measures the information loss using a particular classifier. Because a relatively strong classifier generally captures the predictive information from features better than a weak classifier, the accuracy of a strong classifier may be more consistent with the amount of predictive information contained in a feature subset.

\def\myWidth{2.5}
\begin{figure}[h]
\centering
\includegraphics[width= \myWidth in]{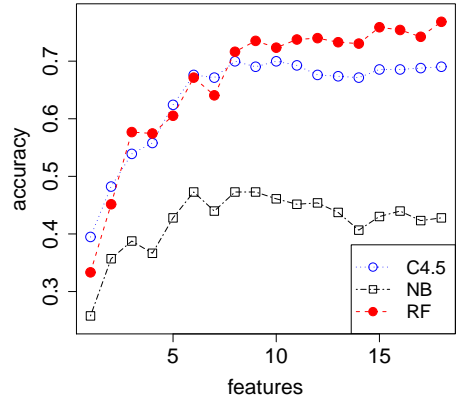} 
\caption{Accuracy of C4.5, naive Bayes (NB) and random forest (RF) for different numbers of features for the Vehicle data set from the UCI database. Starting from an empty feature set, each time a new feature is randomly selected and added to the set. The accuracy of RF continues to improve as more features are used, while the accuracy of C4.5 and NB stops improving after adding a certain number of features. \label{fig:feaDemo}}
\end{figure}

To illustrate this point, we randomly split the Vehicle data set from the UCI database \cite{blake1998uci} into a training set and a testing set with the same number of instances. Starting from an empty feature set, each time a new feature was randomly selected and added to the set. Then C4.5 \cite{quinlan1993c4}, NB, and a relatively strong classifier random forest (RF) \cite{breiman2001} were trained using the feature subsets, respectively. The accuracy of each classifier on the testing set versus the number of features is shown in Figure \ref{fig:feaDemo}. For C4.5 and NB, the accuracy stops increasing after adding a certain number of features. However, RF continues to improve as more features are added, which indicates the added features contain additional predictive information. Therefore, compared to RF, the accuracy performance of C4.5 and NB may be less consistent with the amount of predictive information contained in the features. This point is also validated by 
experiments shown later in this paper. Furthermore, in many cases higher classification accuracy and thus a relatively strong classifier may be preferred. Therefore, a feature selection method capable of producing a high-quality feature subset with regard to a strong classifier is desirable.

\begin{table*}[ht]
\scriptsize
\centering
\begin{tabular}{|c|c|c|c||c|c|c|c|}
\hline
 Data &  instances &   features &    classes &  Data &  instances &   features &    classes \\
\hline
\hline
    german &       1000 &         20 &          2 &        ada &       4147 &         48 &          2 \\
\hline
  waveform &       5000 &         21 &          3 &      sonar &        208 &         60 &          2 \\
\hline
     horse &        368 &         22 &          2 & HillValley &        606 &        100 &          2 \\
\hline
parkinsons &        195 &         22 &          2 &       musk &        476 &        166 &          2 \\
\hline
      auto &        205 &         25 &          6 & arrhythmia &        452 &        279 &         13 \\
\hline
      hypo &       3163 &         25 &          2 &    madelon &       2000 &        500 &          2 \\
\hline
      sick &       2800 &         29 &          2 &       gina &       3153 &        970 &          2 \\
\hline
      iono &        351 &         34 &          2 &       hiva &       3845 &       1617 &          2 \\
\hline
    anneal &        898 &         38 &          5 &     arcene &        100 &      10000 &          2 \\
\hline
\end{tabular}
\caption{Summary of the data sets used in experiments.\label{table:data}}
\end{table*}

\begin{table*}[!]
\scriptsize
\centering
\begin{tabular}{|c|c|c|c|c|c||c|c|c|c|c|c|}
\hline
      Data &        All &        RRF &     RBoost &        CFS &       FCBF &       Data &        All &        RRF &     RBoost &        CFS &       FCBF \\
\hline
\hline
    german &         20 &       17.9 &       18.7 &        4.9 &        3.6 &        ada &         48 &       39.1 &       41.2 &        8.4 &        7.0 \\
\hline
  waveform &         21 &       21.0 &       21.0 &       15.3 &        7.1 &      sonar &         60 &       18.9 &       21.4 &       10.8 &        6.6 \\
\hline
     horse &         22 &       18.4 &       19.3 &        3.9 &        3.9 & HillValley &        100 &       30.7 &       33.5 &        1.0 &        1.0 \\
\hline
parkinsons &         22 &       10.6 &       12.3 &        7.8 &        3.5 &       musk &        166 &       34.5 &       34.8 &       29.2 &       11.0 \\
\hline
      auto &         25 &        8.2 &        8.4 &        6.8 &        4.5 & arrhythmia &        279 &       26.8 &       28.9 &       17.7 &        8.2 \\
\hline
      hypo &         25 &       12.4 &       14.5 &        5.3 &        5.5 &    madelon &        500 &       72.5 &       76.9 &       10.7 &        4.7 \\
\hline
      sick &         29 &       12.3 &       16.3 &        5.4 &        5.6 &       gina &        970 &       83.0 &       95.4 &       51.6 &       16.1 \\
\hline
      iono &         34 &       15.2 &       18.5 &       11.7 &        9.1 &       hiva &       1617 &      146.1 &      192.6 &       38.6 &       13.6 \\
\hline
    anneal &         38 &       11.5 &       11.7 &        5.8 &        6.9 &     arcene &      10000 &       22.5 &       28.2 &       49.4 &       35.1 \\
\hline
\end{tabular}
\caption{The total number of features (``All"), and the average number of features selected by different feature selection methods.
\label{table:numfea}}
\end{table*}

\long\def\symbolfootnote[#1]#2{\begingroup%
\def\thefootnote{\fnsymbol{footnote}}\footnote[#1]{#2}\endgroup}

\begin{table*}[ht]
\scriptsize
\centering
\begin{tabular}{|c|c|cc|cc|cc|cc|c|cc|cc|cc|cc|}
\hline
           &                                                                              \multicolumn{ 9}{|c|}{Classifier: RF} &                                                                            \multicolumn{ 9}{|c|}{Classifier: C4.5} \\
\hline
        Data   &        All & \multicolumn{ 2}{|c|}{RRF} & \multicolumn{ 2}{|c|}{RBoost} & \multicolumn{ 2}{|c|}{CFS} & \multicolumn{ 2}{|c|}{FCBF} &        All & \multicolumn{ 2}{|c|}{RRF} & \multicolumn{ 2}{|c|}{RBoost} & \multicolumn{ 2}{|c|}{CFS} & \multicolumn{ 2}{|c|}{FCBF} \\
\hline
\hline
    german &      0.752 &      0.750 &            &      0.750 &            &      0.704 &       $-$ &      0.684 &       $-$ &      0.716 &      0.719 &            &      0.716 &            &      0.723 &            &      0.713 &            \\
\hline
  waveform &      0.849 &      0.849 &            &      0.849 &            &      0.846 &       $-$ &      0.788 &       $-$ &      0.757 &      0.757 &            &      0.757 &            &      0.765 &       $+$ &      0.749 &       $-$ \\
\hline
     horse &      0.858 &      0.857 &            &      0.853 &       $-$ &      0.824 &       $-$ &      0.825 &       $-$ &      0.843 &      0.843 &            &      0.842 &            &      0.835 &            &      0.836 &            \\
\hline
parkinsons &      0.892 &      0.891 &            &      0.891 &            &      0.878 &       $-$ &      0.846 &       $-$ &      0.842 &      0.843 &            &      0.841 &            &      0.841 &            &      0.839 &            \\
\hline
      auto &      0.756 &      0.756 &            &      0.759 &            &      0.746 &            &      0.715 &       $-$ &      0.662 &      0.634 &            &      0.638 &            &      0.637 &            &      0.640 &            \\
\hline
      hypo &      0.989 &      0.990 &       $+$ &      0.990 &       $+$ &      0.985 &       $-$ &      0.990 &            &      0.992 &      0.992 &            &      0.992 &            &      0.988 &       $-$ &      0.991 &            \\
\hline
      sick &      0.979 &      0.981 &       $+$ &      0.980 &       $+$ &      0.966 &       $-$ &      0.966 &       $-$ &      0.982 &      0.982 &            &      0.982 &            &      0.973 &       $-$ &      0.973 &       $-$ \\
\hline
      iono &      0.931 &      0.926 &            &      0.928 &            &      0.925 &       $-$ &      0.919 &       $-$ &      0.887 &      0.881 &            &      0.881 &            &      0.889 &            &      0.880 &            \\
\hline
    anneal &      0.944 &      0.940 &       $-$ &      0.941 &            &      0.904 &       $-$ &      0.919 &       $-$ &      0.897 &      0.896 &            &      0.893 &            &      0.869 &       $-$ &      0.890 &            \\
\hline
       ada &      0.840 &      0.839 &            &      0.839 &            &      0.823 &       $-$ &      0.831 &       $-$ &      0.830 &      0.829 &            &      0.830 &            &      0.842 &       $+$ &      0.840 &       $+$ \\
\hline
     sonar &      0.803 &      0.783 &       $-$ &      0.774 &       $-$ &      0.739 &       $-$ &      0.734 &       $-$ &      0.701 &      0.693 &            &      0.691 &            &      0.689 &            &      0.697 &            \\
\hline
HillValley &      0.546 &      0.511 &       $-$ &      0.514 &       $-$ &      0.489 &       $-$ &      0.498 &       $-$ &      0.503 &      0.503 &            &      0.503 &            &      0.503 &            &      0.503 &            \\
\hline
      musk &      0.865 &      0.849 &       $-$ &      0.853 &       $-$ &      0.840 &       $-$ &      0.821 &       $-$ &      0.766 &      0.746 &       $-$ &      0.768 &            &      0.771 &            &      0.752 &            \\
\hline
arrhythmia &      0.682 &      0.704 &       $+$ &      0.699 &       $+$ &      0.721 &       $+$ &      0.685 &            &      0.642 &      0.648 &            &      0.649 &            &      0.662 &       $+$ &      0.657 &            \\
\hline
   madelon &      0.671 &      0.706 &       $+$ &      0.675 &            &      0.784 &       $+$ &      0.602 &       $-$ &      0.593 &      0.661 &       $+$ &      0.643 &       $+$ &      0.696 &       $+$ &      0.611 &       $+$ \\
\hline
      gina &      0.924 &      0.915 &       $-$ &      0.914 &       $-$ &      0.891 &       $-$ &      0.832 &       $-$ &      0.847 &      0.851 &            &      0.848 &            &      0.854 &            &      0.817 &       $-$ \\
\hline
      hiva &      0.967 &      0.967 &            &      0.967 &            &      0.966 &            &      0.965 &       $-$ &      0.961 &      0.961 &            &      0.964 &       $+$ &      0.965 &       $+$ &      0.965 &       $+$ \\
\hline
    arcene &      0.760 &      0.683 &       $-$ &      0.676 &       $-$ &      0.713 &       $-$ &      0.702 &       $-$ &      0.603 &      0.633 &            &      0.606 &            &      0.566 &            &      0.586 &            \\
\hline
\hline
{\bf win/lose/tie} &    {\bf -} & \multicolumn{ 2}{|c|}{{\bf 4/6/8}} & \multicolumn{ 2}{|c|}{{\bf 3/6/9}} & \multicolumn{ 2}{|c|}{{\bf 2/14/2}} & \multicolumn{ 2}{|c|}{{\bf 0/16/2}} &    {\bf -} & \multicolumn{ 2}{|c}{{\bf 1/1/16}} & \multicolumn{ 2}{|c}{{\bf 2/0/16}} & \multicolumn{ 2}{|c}{{\bf 5/3/10}} & \multicolumn{ 2}{|c|}{{\bf 3/3/12}} \\
\hline
\end{tabular}
\caption{The average accuracy of random forest (RF) and C4.5 applied to all features, and the feature subsets selected by different methods respectively. The feature subsets having significantly better/worse accuracy than all features at a 0.05 level are denoted as +/-. \label{table:RF}} 
\end{table*}
\section{Experiments}\label{sec:experiment}
Data sets from the UCI benchmark database \cite{blake1998uci}, the NIPS 2003 feature selection benchmark database, and the IJCNN 2007 Agnostic Learning vs. Prior Knowledge Challenge database were used for evaluation. These data sets are summarized in Table \ref{table:data}. We implemented the regularized random forest (RRF) and the regularized boosted random trees (RBoost) under the Weka framework \cite{weka}. Here $\lambda=0.5$ is used and initial experiments show that, for most data sets, the classification accuracy results do not change dramatically with $\lambda$. 

The regularized trees were empirically compared to CFS \cite{hall2000}, FCBF \cite{liuhuan2004}, and SVM-RFE \cite{guyon2002}. These methods were selected for comparison because they are well-recognized and widely-used. These methods were run in Weka with the default settings. 



\def\myWidth{2.8}
\begin{figure*}[ht]
\centering
\subfigure[{The musk data. The SVM-RFE took 109 seconds to run, while RRF took only 4 seconds on average.}]{
\includegraphics[width= \myWidth in]{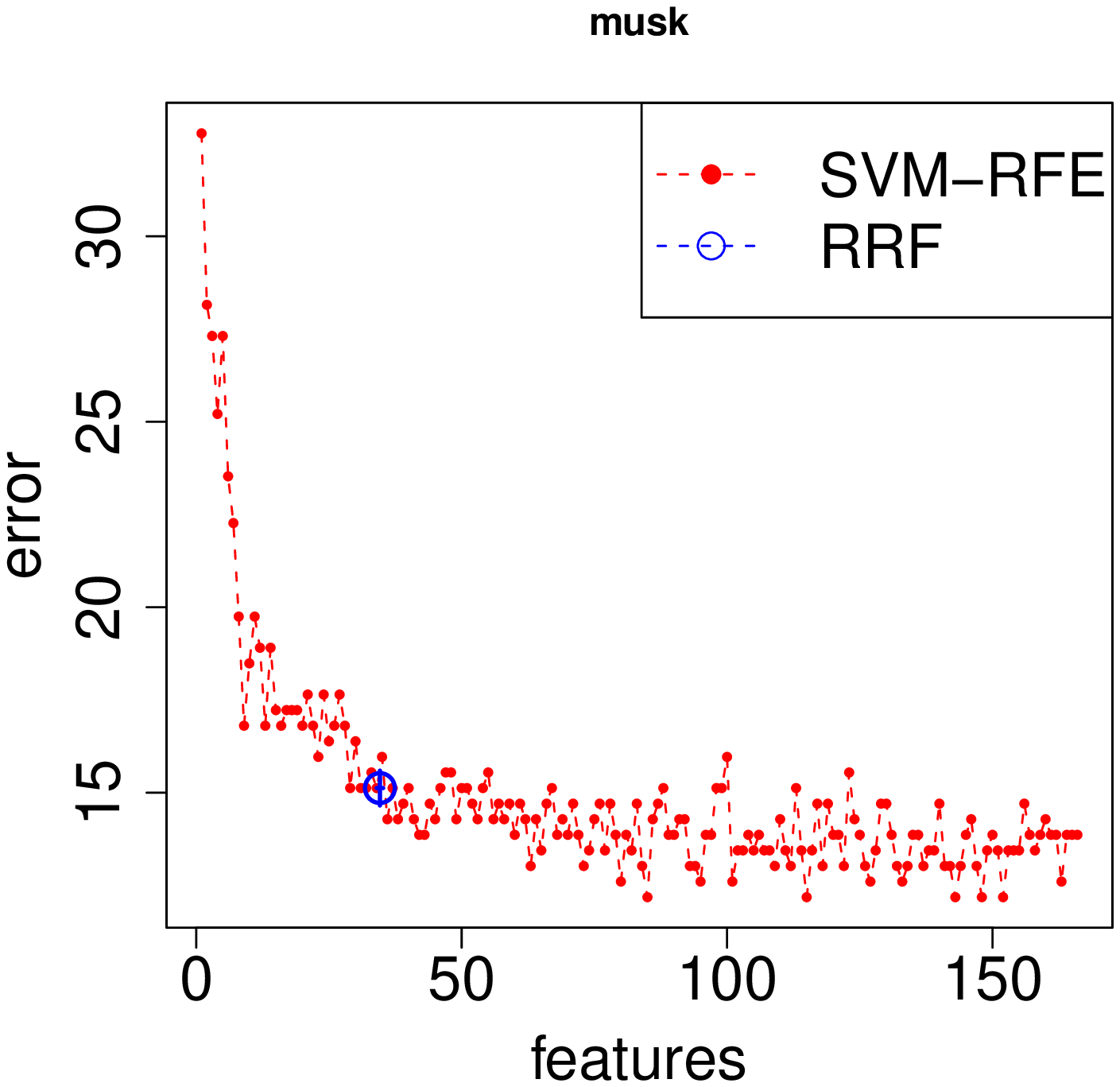} 
}
\subfigure[The arrhythmia data. SVM-RFE took 442 seconds to run, while RRF took only 6 seconds on average.]{
\includegraphics[width= \myWidth in]{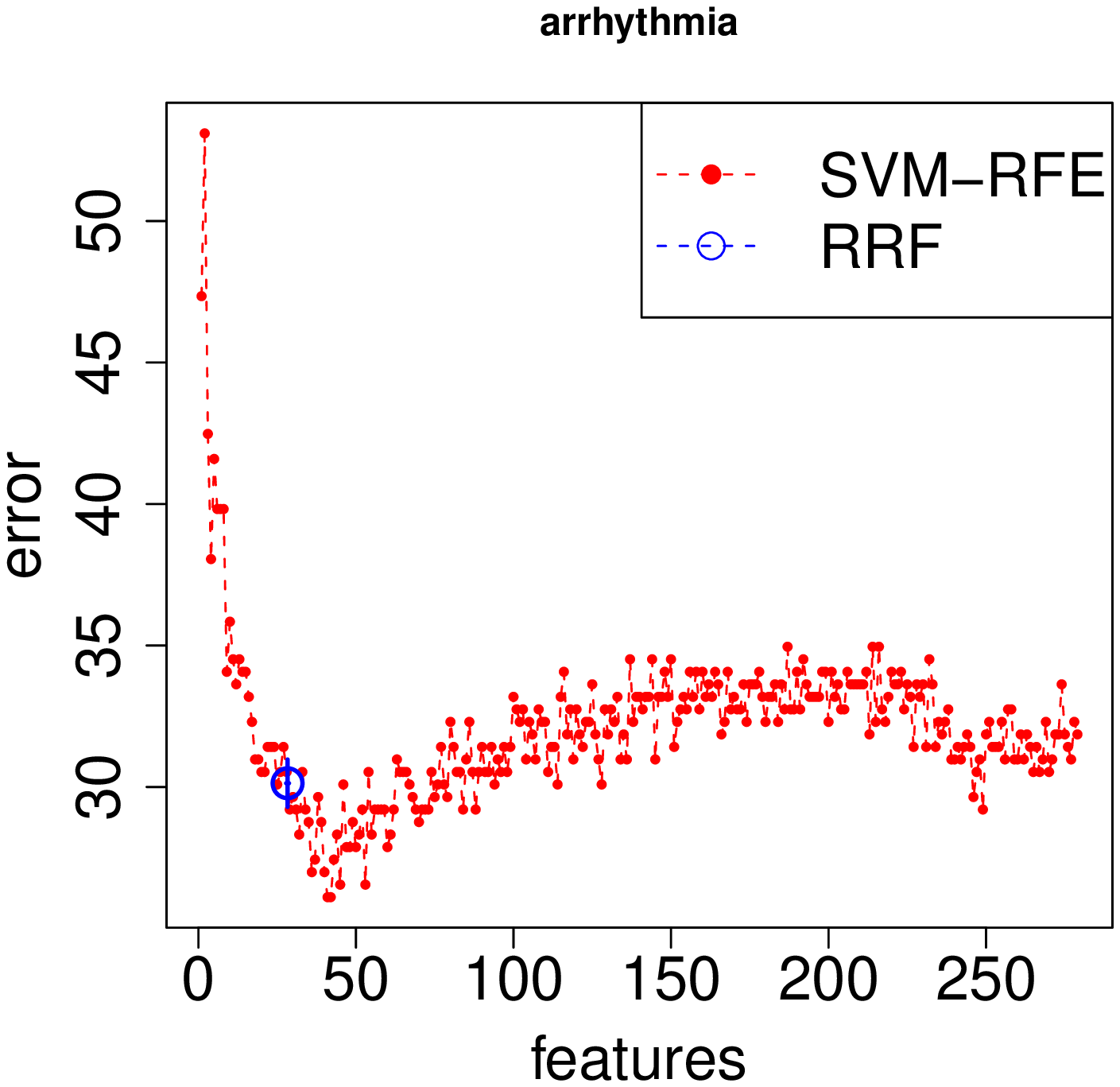} 
}
\caption[2]{The results of SVM-RFE and RRF. Plotted points show the errors versus the number of backward elimination iterations used in SVM-RFE. The circles correspond to the average error versus the average number of features over 10 runs of RRF. The straight lines on the circles are the standard errors (vertical lines) or number of features (horizontal lines). \label{fig:SVMRRF}}
\end{figure*}

We applied the following classifiers: RF (200 trees) \cite{breiman2001} and C4.5 \cite{quinlan1993c4} on all the features and the features selected by RRF, RBoost, CFS and FCBF for each data set, respectively. We ran 10 replicates of two-fold cross-validation for evaluation. Table \ref{table:numfea} shows the number of original features, and the average number of features selected by the different feature selection methods for each data set. Table \ref{table:RF} show the accuracy of RF and C4.5 applied to all features and the feature subsets, respectively. The average accuracy of different algorithms, and a paired t-test between using the feature subsets and using all features over the 10 replicates are shown in the table. The feature subsets having significantly better/worse accuracy than all features at a 0.05 level are denoted as +/-, respectively. The numbers of significant wins/losses/ties using the feature subsets over using all features are also shown.

Some trends are evident. In general, CFS and FCBF tend to select fewer features than the regularized tree ensembles. However, RF using the features selected by CFS or FCBF has many more losses than wins on accuracy, compared to using all the features. Note both CFS and FCBF consider only two-way interactions between the features, and, therefore, they may miss some features which are useful only when other features are present. In contrast, RF using the features selected by the regularized tree ensembles is competitive to using all the features. This indicates that though the regularized tree ensembles select more features than CFS and FCBF, these additional features indeed add additional predictive information. For some data sets where the number of instances is small (e.g. arcene), RF using the features from RRF or RBoost do not have an advantage over RF using the features from CFS. This may be because a small number of instances leads to small trees, which are less capable of capturing multi-way feature interactions.

The relatively weak classifier C4.5 performs differently from RF. The accuracy of C4.5 using the features from every feature selection method is competitive to using all the features, even though the performance of RF suggests that CFS and FCBF may miss some useful predictive information. This indicates that that C4.5 may be less capable than RF on extracting predictive information from features.

In addition, the regularized tree ensembles: RRF and RBoost have similar performances regarding the number of features selected or the classification accuracy over these data sets.

Next we compare the regularized tree ensembles to SVM-RFE. For simplicity, here we only compare RRF to SVM-RFE. The algorithms are evaluated using the musk and arrhythmia data sets. Each data set is split into the training set and testing set with equal number of instances. The training set is used for feature selection and training a RF classifier, and the testing set is used for testing the accuracy of the RF. Figure \ref{fig:SVMRRF} plots the RF accuracy versus the number of backward elimination iterations used in SVM-RFE.
Note that RRF can automatically decide the number of features. Therefore, the accuracy of RF using the features from RRF is a single point on the figure. We also considered the randomness of RRF. We run RRF 10 times for each data set and Figure \ref{fig:SVMRRF} shows the average RF error versus the average number of selected features. The standard errors are also shown. 

For both data sets, RF's accuracy using the features from RRF is competitive to using the optimum point of SVM-RFE. It should be noted that SVM-RFE still needs to select a cutoff value for the number of features by strategies such as cross-validation, which not necessarily selects the optimum point, and \textsl{also} further increase the computational time. Furthermore, RRF (took less than 10 seconds in average to run for each data set) is considerably more efficient than SVM-RFE (took more than 100 seconds to run for each data set).  

\section{Conclusion}\label{sec:conclusion}
We propose a tree regularization framework, which adds a feature selection capability to many tree models. We applied the regularization framework on random forest and boosted trees to generate regularized versions (RRF and RBoost, respectively). Experimental studies show that RRF and RBoost produce high-quality feature subsets for both strong and weak classifiers. As tree models are computationally fast and can naturally deal with categorical and numerical variables, missing values, different scales (units) between variables, interactions and nonlinearities etc., the tree regularization framework provides an effective and efficient feature selection solution for many practical problems.

\section*{Acknowledgements}
This research was partially supported by ONR grant N00014-09-1-0656.

\bibliographystyle{IEEEtran}

\bibliography{LTree}
\end{document}